\pdfoutput=1

\documentclass[11pt]{article}

\usepackage[]{acl}

\usepackage{times}
\usepackage{latexsym}
\usepackage{graphicx}
\usepackage{caption}
\usepackage{subcaption}

\usepackage[T1]{fontenc}
\usepackage{booktabs}

\usepackage[utf8]{inputenc}

\usepackage{microtype}
\usepackage{makecell}
\usepackage{xcolor}

\newcommand{\ra}[1]{\renewcommand{\arraystretch}{#1}}
\newcolumntype{L}[1]{>{\raggedright\arraybackslash}p{#1}}
\title{Better Quality Estimation for Low Resource Corpus Mining}

\author{Muhammed Yusuf Kocyigit \\
  Boston University \\
  \texttt{kocyigit@bu.edu} \\\And
  Jiho Lee \\
  Boston University \\
  \texttt{jiholee@bu.edu} \\\And
  Derry Wijaya \\
  Boston University \\
  \texttt{wijaya@bu.edu} \\}
\begin{document}
\maketitle
\begin{abstract}
Quality Estimation (QE) models have the potential to change how we evaluate and maybe even train machine translation models. However, these models still lack the robustness to achieve general adoption. We show that State-of-the-art QE models, when tested in a Parallel Corpus Mining (PCM) setting, perform unexpectedly bad due to a lack of robustness to out-of-domain examples. We propose a combination of multitask training, data augmentation and contrastive learning to achieve better and more robust QE performance. We show that our method improves QE performance significantly in the MLQE challenge and the robustness of QE models when tested in the Parallel Corpus Mining setup. We increase the accuracy in PCM by more than $0.80$, making it on par with state-of-the-art PCM methods that use millions of sentence pairs to train their models. In comparison, we use \emph{thousand} times less data, 7K parallel sentences in total, and propose a novel low resource PCM method.
\end{abstract}

\section{Introduction}

The Quality Estimation (QE) task aims to model human perception of translation quality and predict the quality score an expert would give to a translation using only the source sentence and the translation. This requires the QE model to represent the cross-lingual similarity between source and hypothesis sentences while incorporating different features of the hypothesis sentence such as fluency, grammaticality and adequacy\footnote{Fluency
measures whether a translation is fluent, regard-
less of the correct meaning, while Adequacy measures whether the translation conveys the correct
meaning, even if the translation is not fully fluent \cite{snover2009fluency}}.

Human evaluations of machine translation are costly and time-consuming for a large-scale text dataset. References to evaluate machine translation performance are not readily available in many cases, especially in low-resource languages. Even if they do exist, they often assume a single, unique answer for correct translations, causing bias in the evaluation. Thus, it is academically and professionally of paramount importance to further develop reliable Quality Estimation metrics, which can ultimately eliminate the need for references and have unlimited potential for practical applications in machine translations. 

Parallel Corpus Mining (PCM) is another critical task that can enable the creation of high-quality parallel data and reduce the need for considerable human effort. These mined parallel corpora could especially be helpful in low resource languages. On the other hand, current PCM methods require large amounts of parallel data, which creates a paradoxical loop that only large companies can break.

Quality estimation is uniquely linked with PCM since what makes a good translation most of the time makes a correct parallel too. Considering the similarity in the underlying goals of these two tasks, we expect models that can do one to perform, at least, acceptably in the other. However, \citet{Zhao2020OnTL} have shown that models that can do corpus mining fail in QE and propose a resource prudent method to bridge the gap. We show that the gap exists in the other direction and we introduce simple and valuable solutions. 

In chapters \ref{sect:quality estimation} and \ref{sect:quality experiments}, we introduce our method MultiQE and its base variants. Since we do not want to depend on additional cross-lingual data, we propose using multitask training with monolingual linguistic inference and semantic similarity data. We also experiment with using multitask feature extraction and compare our methods with SoTA QE methods in the Multilingual Quality Estimate (MLQE) dataset \cite{tacl2020}. 

In chapter \ref{sect:pcm}, we use data augmentation techniques in combination with multitask training to train more robust QE models and check their robustness in the Parallel Corpus Mining setup using the TATOEBA\cite{tatoeba} and BUCC\cite{bucc2018} datasets. We use the term robust QE models to refer to models that can overcome the problem of just focusing on grammaticality/fluency, which causes SoTA QE models to fail in PCM. Our method outperforms SoTA QE models on PCM with a substantial margin, up to $0.80$ difference in accuracy score in TATOEBA. 

In addition, we compare our method with high resource methods like LASER \cite{Artetxe2019MassivelyMS} and LaBSE \cite{LaBSE} which are trained on vast amounts of parallel data and achieve SoTA performances on PCM. Our method essentially offers a better and more robust QE model that is trained with very little data (thousand times less data) compared to these models. The goal in comparing to these high resource methods is to show that our proposed method achieves good enough performance to be a viable \emph{low resource} PCM method. Our contributions in this paper can be summarized as below.
\begin{itemize}
    \item We propose using multitask training for QE with STS (Semantic Textual Similarity) and MNLI (Multi-Genre Natural Language Inference) and show that even though these datasets are monolingual, multitask training can improve QE performance in MLQE significantly. 
    \item We propose a robustness test for QE models through the PCM setting showing that SoTA QE models fail this test. We test how our multitask training method performs and propose using negative data augmentation to improve robustness further. We demonstrate that multitask training and negative data augmentation improve the robustness of QE models with an 0.80 increase in accuracy in the TATOEBA challenge.
    \item We propose a viable low resource corpus mining approach involving a sentence embedding model trained with the contrastive loss on the QE dataset and our robust QE model. We show that our method performs better under low resource conditions and is even comparable in high resource settings to SoTA in Parallel Corpus Mining.  

\end{itemize}

\section{Related Work}
\subsection{Quality Estimation}
\textbf{State of the art in QE} In sentence-level Quality Estimation, multilingual language models as well as machine translation models are used for getting sentence representations as features to train quality estimation models \citep{yankovskaya-etal-2019-quality}, \citep{kim2017predictor},  \citep{zhou-etal-2019-source}  \citep{Peters2018DeepCW}. Similarly TransQuest \cite{transquest:2020} uses a cross-lingual transformer language model, XLM-R \citep{xlmrconneau}, to extract features for sentence-level Direct Assessment scores and achieves SoTA performance in WMT-2020 QE task. This MonoTransQuest architecture will be used as our baseline.

\textbf{Multitask Learning in QE} Multitask learning is shown to be effective for QE. \citet{kim-etal-2019-qe} create a combined loss focusing on all QE tasks at once. They train a bilingual BERT to extract sentence representations. This model simultaneously predicts word quality tags(GOOD or BAD from the word level QE task) HTER score and takes the last hidden layer as the features for sentence level QE. They limit their work to signals from the MLQE dataset's word and sentence level tasks and do not apply to external datasets, unlike our work.

\textbf{External Signals in QE} \citet{lo-2019-yisi} enhance their embeddings with semantic role labels and show that it improves QE performance, demonstrating the importance of semantic features in QE. \citet{martins-etal-2017-pushing} use part of speech tagging and show that it can also improve the QE performance. 

\textbf{Usage of NLI and STS} Pretraining the backbone via multitask training, using NLI and STS, has been shown to improve performance in translation evaluation with references. By allowing the backbone network to learn the cross relations between sentences from different aspects, \citet{bleurt} use this framework by including the linguistic inference task and achieve SoTA performance on machine translation evaluation with references. Another method that performs comparably is \cite{kanenubia}, where the authors use separately pre-trained models to extract features and later train a final layer to evaluate translations with references. 

\subsection{Cross-Lingual Alignment} 
\label{rel:crosslingalignment}
\textbf{Motivation for Alignment} \citet{Zhao2020OnTL} find that cross-lingual encoders such as XLM \citep{lample2019cross} and M-BERT make mistakes in QE. They realize that the same sentence in different languages are not close to each other in the multilingual embedding space due to changing sentence structures, which they call semantic mismatch. \citet{Zhao2020OnTL} show that aligned embeddings perform much better than directly using the backbone. Since we want to benefit from monolingual datasets, we wanted to check how aligned feature extractors would fare against regular multitask training and the current SoTA in QE.

\textbf{Motivation for Translation} Recent work has shown that in some cases, translating one of the sentences can also work just as well as alignment \citep{Conneau2018XNLIEC}. Hence we also compare translating one of the sentences to aligning the representations of the non-English sentences from the XLM-R model similar to \citet{Conneau2018XNLIEC}. 

\subsection{Parallel Corpus Mining}\
\textbf{State of the art in PCM} For Parallel Corpus Mining, models are generally trained on large parallel corpora. \citet{Artetxe2019MassivelyMS} train an encoder-decoder network on large scale translation data and use the encoder output as an embedding space to compare sentences. \citet{MUSE} train a network on the translation ranking problem, sampling a number of negative examples from the corpus for each input sentence.

\textbf{Motivation for using QE in PCM} \citet{Reimers2020MakingMS} train a cross-lingual language model(student) to imitate the embedding space of another sentence embedding model(teacher) trained on a related task like paraphrase detection, STS or NLI. They show that the usage of external tasks can improve performance in PCM. Although their method is remarkable, it still requires a lot of parallel data to align the XLM-R with the embeddings of the new network. We also observe that alignment under low resource conditions is not very effective. During our experiments, we looked into viable ways of using QE data for training models to perform well in PCM with low resource limitations in mind. Since all these methods use a large amount of parallel data from a variety of sources and datasets, introducing a method that can achieve similar scores with very little data is an important goal to achieve.

\section{Quality Estimation}
\label{sect:quality estimation}

\subsection{Method}

We compare three different approaches to incorporating STS and NLI tasks into QE. The first one is direct multitask training. The second and third methods use pretraining separate backbone architectures on these tasks and using them to extract features. Because the STS and NLI backbones are trained on monolingual data, we either use cross-lingual alignment of sentence embeddings or translate the non-English sentence to English.

\subsubsection{Multitask Training} \label{multitasktraining}

\begin{figure*}
     \centering
     \resizebox{1\textwidth}{!}{%
     \begin{subfigure}[t]{8cm}
         \centering
         \includegraphics[width=8cm]{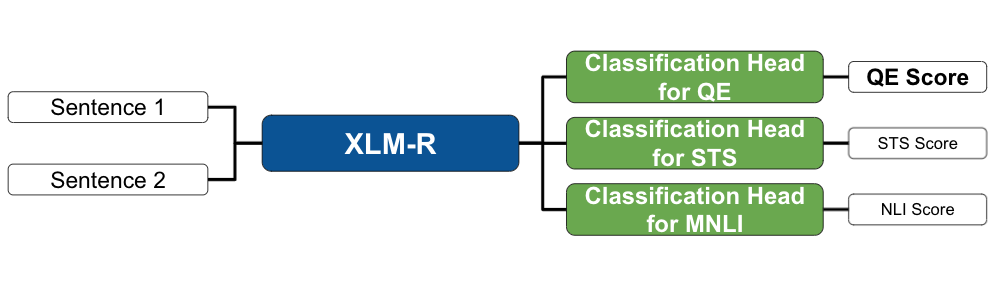}
         \caption{Multitask pretraining for QE: MultiQE Multitask. The three classification heads share the same backbone and the backbone weights are trained during all three phases. Only the head for QE is used after training for obtaining the QE score. Here the classification heads are only a linear layer on top of the mean pooled output}
         \label{fig:multitask}
     \end{subfigure}
     \hspace{4pt}
     \begin{subfigure}[t]{8cm}
         \centering
    \includegraphics[width=8cm]{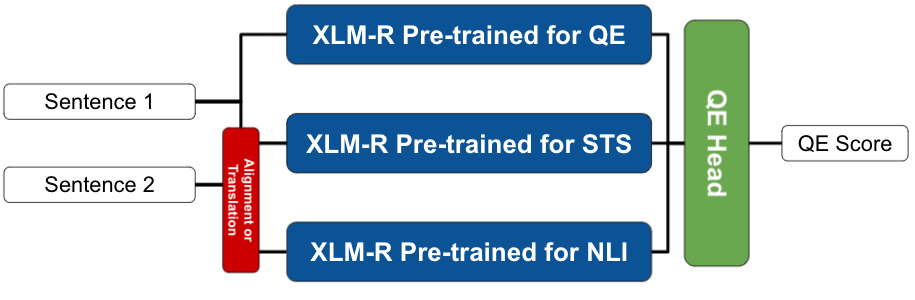}
    \caption{Multitask feature extraction for QE: MultiQE Alignment/Translation. All three backbones are pre-trained on the respective tasks with classification heads on top. The outputs of the backbones are mean pooled to create sentence features. The final QE Head is a two layer fully connected network that is trained on the MLQE dataset.}
    \label{fig:multi_feature}
     \end{subfigure}
     }
    \caption{MultiQE Models.}
    \vspace{-3pt}
\end{figure*}

In this method, we train a single backbone XLM-R model with three classification heads on the STS-B, MNLI and QE tasks. This model will be referred to as MultiQE Multitask. By not doing any explicit alignment, we test if the XLM-R model trained for a cross-lingual task (QE) will benefit from multitask training that includes monolingual data. 
In Figure \ref{fig:multitask}, we illustrate the structure of the multitask learning framework. 

The model is first trained for three epochs and, later, only the quality estimation head with the backbone is fine-tuned for another epoch on QE following insights from \citet{bleurt} 

\begin{table*}\centering
\ra{1}
\resizebox{1\textwidth}{!}{%
\begin{tabular}{@{}llll@{}}\toprule
\textbf{Dataset} & \textbf{Size} & \textbf{Language Pairs}& \textbf{Usage} \\ 
     \midrule
    TATOEBA & <1K & en-de, en-zh, ne-en, si-en& To test performance on Parallel Corpus Mining  \\ 
     \hline
     BUCC & <8k& en-de, en-zh& To test performance on Parallel Corpus Mining\\ 
     \hline
     MLQE  & 7K(Train) 1K(Test) &en-de,en-zh,ro-en,et-en,ne-en,si-en& To train all MultiQE models and test them for QE performance  \\ 
     \hline
     OPUS(JW \& GNOME) & 25K(Train) 3K(Test) & en-de,en-zh,ro-en,et-en,ne-en,si-en& To train and test the alignment module in MultiQE Alignment   \\ 
\bottomrule
\end{tabular}
}
\caption{Parallel datasets, their sizes and how they are used in our methods. The MLQE dataset is created by employing annotators on outputs of machine translation models on the corresponding language. The sentence pairs are labeled on the quality of the translation.}
\label{table:data}
\end{table*}

\subsubsection{Multitask Feature Extraction}

We train three backbones on the STS-B, MNLI and QE datasets and use the extracted features from these models to train a final layer for predicting QE scores. For this model, we compare two approaches: the first one is named MultiQE Alignment and is explained in section \ref{multiqe:alignment}; the second one is MultiQE Translation, where, instead of aligning sentence embeddings, we translate the non-English sentence to English with Google Translate before inputting the sentence pairs to STS and NLI backbones. In Figure \ref{fig:multi_feature}, we show the general architecture of MultiQE Alignment and Translation. In this architecture, translation and alignment are not used simultaneously. When we use translation, the alignment part is not used and vice versa. 

\subsubsection{Cross-lingual Alignment model}
\label{multiqe:alignment}
We chose to tackle the semantic mismatch problem with a cosine similarity based sentence alignment similar to \citet{Conneau2018XNLIEC}. This alignment pushes the sentence embeddings of the sentences in the non-English languages towards the embeddings of the translations of those sentences in English.

We align the STS and NLI input feature extractors in MultiQE Alignment, using data we get from OPUS  with the cosine similarity objective in Equation \ref{eq:cosine}. Given a set of parallel sentences $X = \{(x_i,y_i)\,\mid\,i=1,2,...,K\}$, we fine tune the model to minimize $L_A$ in Equation \ref{eq:cosine}. 

\begin{equation}
    L_A(X) = \sum_{(x_i,y_i) \in X}(1- \cos(x_{i},y_{i}))
    \label{eq:cosine}
\end{equation}

We test the effectiveness of the alignment using the 3K test sentences we have put aside and measure the cosine similarity before and after alignment, which increases on average from 0.64 to 0.96. 

\section{Quality Estimation Experiments}
\label{sect:quality experiments}
This section will go over the dataset, results, significance test, and ablation study for our experiments on the MLQE dataset.

\begin{table*}\centering
\ra{1}
\resizebox{0.9\textwidth}{!}{%
\begin{tabular}{@{}lcccccc@{}}\toprule
\textbf{Models} & \textbf{en-de} & \textbf{en-zh}& \textbf{ro-en}
     & \textbf{et-en}& \textbf{ne-en}& \textbf{si-en} \\ 
     \midrule
    OpenKiwi \cite{kepler-etal-2019-openkiwi}& 0.145& 0.190& 0.684& 0.477& 0.386& 0.373  \\ 
     MonoTransQuest* \cite{transquest:2020} & 0.408& 0.471& 0.881& 0.754& 0.769& 0.634  \\ 
     MultiQE Translation(Ours) & 0.406& 0.486& \textbf{0.889}&\textbf{0.762}& 0.767& 0.665  \\ 
     MultiQE Alignment(Ours) & 0.415& 0.483& 0.881& 0.756& 0.772& 0.656  \\ 
     MultiQE Multitask(Ours) & \textbf{0.418} &  \textbf{0.512}& 0.879& 0.755&  \textbf{0.777}&  \textbf{0.675} \\
\bottomrule
\end{tabular}
}
\caption{Pearson Correlation with Human Judgment. We observe that multitask training gives the best performance in 4 out of 6 language pairs, while for the mid-resource languages translating the non-English sentence outperforms other methods. We can infer that QE performance can be improved with monolingual NLI and STS data. *Results are reproduced using the Transquest pre-trained model zoo and testing scripts.}
\label{table:corr}
\end{table*}

\subsection{Datasets}
We used the Semantic Textual Similarity - Benchmark(STS-B) dataset for the STS task. This task measures the degree of meaning similarity between sentences with a score ranging from 1-5. STS-B is a collection of English sentence pairs extracted from different publicly available sources.\cite{cer} \cite{glue}
 
For the natural language inference tasks, we use the Multi-Genre Natural Language Inference (MNLI) dataset. The MNLI dataset includes both written and spoken text from various sources. \cite{williams-etal-2018-broad}. The task is to predict the label of entailment, neutral, or contradiction based on a premise and a hypothesis text.

For training and testing on the QE task, we use the Multilingual Quality Estimation (MLQE) dataset, which is derived chiefly from Wikipedia articles \cite{Fomicheva2020UnsupervisedQE}  and contains language pairs from high (en-de, en-zh), medium (ro-en, et-en), and low (ne-en, si-en) resource languages. Each pair has human labels for 7K train, 1K validation and 1K test translation pairs. Quality scores are collected by showing source sentences with translations to 3 experts and averaging the normalized scores.
  
For the cross-lingual alignment  (section \ref{multiqe:alignment}), we use sentence pairs from the JW\cite{agic-vulic-2019-jw300} and GNOME\cite{TIEDEMANN12.463} dataset. We use a small subset(25K) to do the alignment and 3K sentences to test the quality of the alignment, taking low resource conditions into account.

\subsection{Results}
We evaluate our models on the MLQE test set and use Pearson Correlation with human judgment as our primary measure. The results of our methods are in Table \ref{table:corr}. We include \cite{kepler-etal-2019-openkiwi} because it was used as the baseline in the WMT2020 QE challenge. MonoTransQuest is included because it achieves SoTA performance in QE and is the winning entry of the 2020 WMT QE challenge. We use the MonoTransQuest model with no ensemble to have a meaningful comparison.

In Table \ref{table:corr}, we find that multitask training (MultiQE Multitask) and translation (MultiQE Translation) outperform SoTA on all of the language pairs with MultiQE Multitask leading in 4 out of the 6 language pairs. Comparing MultiQE Alignment and MultiQE Translation with MonoTransQuest, all our methods are comparable with previous SoTA if not better.

Among our methods, MultiQE Multitask performs better and is more computationally efficient than MultiQE Alignment and Translation. Since the alignment and translation methods use multiple backbones, they require more computational power in training and inference.


\subsubsection{William's Test} 
\begin{figure*}
     \centering
     \resizebox{0.8\textwidth}{!}{%
     \begin{subfigure}[b]{0.45\textwidth}
         \centering
         \includegraphics[scale = 0.35]{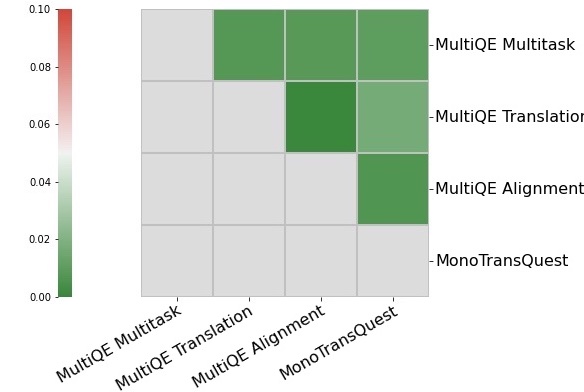}
         \caption{P-values}
         \label{fig:pval}
     \end{subfigure}
     \hfill
     \begin{subfigure}[b]{0.45\textwidth}
         \centering
         \includegraphics[scale = 0.35]{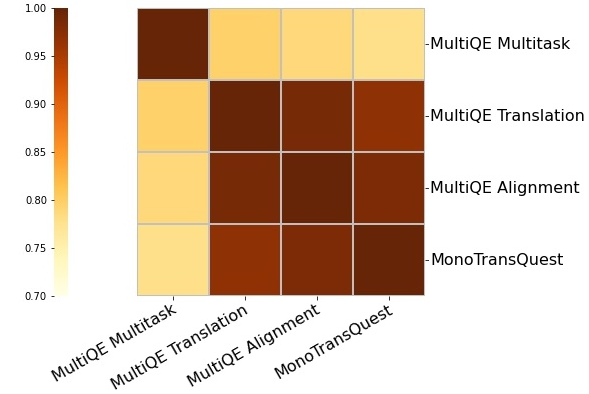}
         \caption{Correlation}
         \label{fig:intercor}
     \end{subfigure}
     }
        \caption{P-values for the Williams Test and Correlation between model predictions. Note that all MultiQE models outperform MonoTransQuest significantly with $p \leq 0.05$. Additionally, we observe that the three methods in the bottom three rows correlate highly while MultiQE Multitask's behavior is different.}
        \vspace{-4mm}
        \label{fig:williams}
\end{figure*}

Correlation scores by themselves are not enough to make conclusions. Therefore, we perform a William's test to check the significance and the inter-correlation between the outputs of the methods. The William's test is performed with the language pair \textit{en-zh}. If we look at Figure \ref{fig:pval}, the P-values are below 0.05, suggesting that our increases in correlation are statistically significant. 

In Figure \ref{fig:intercor}, we find that MultiQE Translation, Alignment, and MonoTransQuest models correlate highly with each other, while MultiQE Multitask can be separated from the others. We would expect a certain level of correlation among these methods because they are run on the same task. However, the high correlation among the first three methods is mainly due to their shared pre-trained backbones.

\subsection{Ablation Study}

Given that the MultiQE Multitask model gives the best performance in QE, we perform the ablation study on this model. The results below (Table \ref{table:abmulti}) are for the \textit{en-zh} language pair. The scores represent Pearson Correlation with Human Judgment. The ablation study explores the effect of these datasets in the pretraining stage. Hence, it does not take out the final QE fine-tuning. Looking at Table \ref{table:abmulti} we observe that STS helps the performance more than MNLI.

\begin{table}\centering
\ra{1}
\resizebox{0.35\textwidth}{!}{%
\begin{tabular}{@{}lc@{}}\toprule
\textbf{Models} & \textbf{en-zh} \\ 
     \midrule
    Multitask MNLI & 0.444  \\ 
    Multitask STS & 0.456  \\ 
    Multitask QE + MNLI & 0.485  \\ 
    Multitask QE + STS & 0.495  \\ 
    Multitask MNLI + STS & 0.471  \\ 
    Multitask QE + MNLI + STS & 0.512  \\ 
\bottomrule
\end{tabular}
}
\caption{Ablation study for the multitask pretraining step of MultiQE Multitask. We observe that the STS dataset improves QE performance more than the MNLI dataset.}
\label{table:abmulti}
\end{table}
 
\section{Parallel Corpus Mining Experiments}
\label{sect:pcm}

In the PCM experiments, we will use MultiQE Multitask because it performs the best in Table \ref{table:corr}.

\subsection{Motivation}
The initial motivation behind testing QE models on PCM sparked from the observation that QE models sometimes assign scores close to 1 to hypothesis sentences that are simple and correct even if they are entirely unrelated to the reference sentence. A sentence like \textit{'December 14, 1964'} would get a high score with many references, most likely because they were never translated wrong and never received a bad score. Stemming from this observation, we wanted a natural setting where we could subject QE models to various sentence pairs and see if they were failing in a particular manner and if we could remedy this. Corpus mining was a good candidate because we would have to check every hypothesis sentence for each reference creating a variety of pairs and we would also have the gold labels for correct pairs. Essentially we used the PCM setup as a stress test for QE models. Observing how QE models failed this test and through solving both the computation and performance problems, we have introduced a novel low resource corpus mining method based on the QE task.

\begin{table*}\centering
\ra{1}
\resizebox{1\textwidth}{!}{%
\begin{tabular}{@{}lclcclcclccl@{}}\toprule
& \multicolumn{2}{c}{\textbf{en-de}} & \phantom{abc}& \multicolumn{2}{c}{\textbf{en-zh}} &
\phantom{abc} & \multicolumn{2}{c}{\textbf{ne-en}}& \phantom{abc}&\multicolumn{2}{c}{\textbf{si-en}}\\
\cmidrule{2-3} \cmidrule{5-6} \cmidrule{8-9} \cmidrule{11-12}
& Score & Datasize & & Score & Datasize & & Score & Datasize & & Score & Datasize \\ \midrule
MonoTransQuest \cite{transquest:2020} & 0.07 & \textbf{7K} & &0.05 & \textbf{7K} & &0.12 & \textbf{7K} & &0.20 & \textbf{7K}  \\
LASER \cite{Artetxe2019MassivelyMS}  & \textbf{0.99} & 8.7M & & \textbf{0.95} & 8.3M & & 0.38$^*$ & 0 & & 0.55 &  796K  \\ 
LaBSE \cite{LaBSE} & 0.97 & 100M & &\textbf{0.95} & 100M & &0.85 & 20M+ & &\textbf{0.92} & 20M+ \\ 
Knowledge Distillation \cite{Reimers2020MakingMS} &0.97 & 25M & &0.94 & 12M+ & & 0.41$^*$ & 0 & & 0.12$^*$ & 0  \\ 
MultiQE Multitask (Ours) & 0.03 & \textbf{7K} & & 0.64 & \textbf{7K} & &0.53 & \textbf{7K} & &0.46 & \textbf{7K}   \\ 
MultiQE Multitask + DA (Ours) & 0.97 & \textbf{7K} & &\textbf{0.95} & \textbf{7K} & & \textbf{0.86} & \textbf{7K} & &0.74 & \textbf{7K} \\
\bottomrule
\end{tabular}
}
\caption{Accuracy for the \textit{TATOEBA:Similarity Search Challenge} and the amount of parallel data used by that model for that language pair. Our method achieves SoTA performance in 2 out of the 4 language pairs while it is also comparable in en-de. Our method also outperforms LASER on \textit{si-en} where this method has an order of magnitude closer data with our method. This is especially interesting since it strengthens the argument that our method performs better in low resource regimes. $^*$ signifies that the method does not have support for that language pair, but they can have access to data for similar languages.}
\label{table:tatoeba}
\end{table*}

\subsection{Datasets}
We evaluate our models for PCM on the BUCC \cite{bucc2018} and TATOEBA \cite{tatoeba} datasets. In the BUCC challenge, the goal is to extract ground-truth parallel sentences that are injected into relevant Wikipedia articles. The injected parallel sentences come from the News Commentary Dataset \cite{TIEDEMANN12.463}. The performance is evaluated with the F1 score. In the TATOEBA challenge, the task is to find the translation for each sentence. The TATOEBA challenge contains translation pairs from various sources of more than 100 language pairs. For high resource language pairs, we use the 1000 sentence test set from LASER repository\footnote{\href{https://github.com/facebookresearch/LASER}{https://github.com/facebookresearch/LASER}} because the methods we compare to \citep{Reimers2020MakingMS, Artetxe2019MassivelyMS, LaBSE} also used this test set. However, for low resource languages that are not present in the LASER repository, we use the TATOEBA (2021-08-07) dataset.

\subsection{Method}
Here we will introduce our negative data augmentation scheme and how we offer to solve the computational cost problem by training a sentence embedding model with contrastive loss.
\subsubsection{Model Training}
For parallel corpus mining, we use a scoring model, MultiQE Multitask, and a filtration model. The scoring model takes sentence pairs as inputs and when the size of the corpus to mine gets larger, the cost for computing scores of all sentence pairs gets too high as explained in \citet{reimers-2019-sentence-bert}. To tackle this, we train a sentence embedding model to do pre-filtration of the raw corpus to reduce the search space to a reasonable size. The filtration model is only used for pre-filtration and not the final sentence pair scoring. For small datasets, the scoring model can be used alone. 

The sentence filtration model is trained on the MLQE training data using a contrastive loss. For a set of sentence pairs $I =\{(x_i,z_i)\,\mid\,i=1,2,....,N\}$, we sample a subset of $n$ negative samples for each $x_i$ to form the set $\hat{I}$ such that $\hat{I} = \{(x_i,{z_j}_{j \neq i})\,\mid\,i,j=1,2,....,N\}$. Here, we choose $n$ to be 3, exclude samples from $I$ that have a lower quality score than 0.7, and include them in $\hat{I}$. The labels, $Y_{F}$ for filtration, for each pair in set $I$ are 1 and the labels for each pair in $\hat{I}$ are 0. The filtration model is later trained on the two sets using the loss function given in Equation \ref{eq:contrastive} from \citet{lecun}: 
\begin{equation} 
\begin{array}{l}
    L_F(I,Y_F) = \\ 
    (1-Y_F)\frac{1}{2}D(I)^2 + (Y_F)\frac{1}{2}\{\max(0,m-D(I)\}^2 
    \label{eq:contrastive}
\end{array}
\end{equation}

$D(I)$ represents the similarity metric given a set of sentence pairs $I$ and the subscript F denotes that the labels and loss are for the filtration model. We calculate $D(I)$ as the cosine similarity between the embeddings $(G(x_i),G(z_i))$ of the two sentences $(x_i,z_i)$ where $G$ is the embedding network

\begin{equation} 
\begin{array}{l}
    D(I) = \frac{\vec{G(x_i)} \cdot \vec{G(z_i)}}{\|G(x_i)\| \|G(z_i)\|}
    \label{eq:cosine2}
\end{array}
\end{equation}

\begin{table*}\centering
\ra{1}
\resizebox{1\textwidth}{!}{%
\begin{tabular}{@{}lclcclccl@{}}\toprule
& \multicolumn{2}{c}{en-de} & \phantom{abc}& \multicolumn{2}{c}{en-zh} &
\phantom{abc} & \multicolumn{2}{c}{Average}\\
\cmidrule{2-3} \cmidrule{5-6} \cmidrule{8-9}
& Score & Datasize & & Score & Datasize & & Score & Datasize \\ \midrule
mUSE \cite{MUSE} & 88.5 & 60M+ &  & 86.9 & 60M+ & & 87.7 & 60M+  \\
LASER \cite{Artetxe2019MassivelyMS} & 95.4 & 8.7M & & 91.7 & 8.3M & & 93.5 & 8.4M  \\ 
LaBSE \cite{LaBSE}& \textbf{95.9} & 100M & &\textbf{93.0} & 100M & &\textbf{94.4} & 100M  \\ 
Knowledge Distillation \cite{Reimers2020MakingMS} & 90.8 & 25M & & 87.8 & 12M+ & & 89.3 & 18M+ \\ 
MultiQE Multitask + DA (Ours) & 85.4 & 7K & &75.1 & 7K & &80.2 & 7K\\
\bottomrule
\end{tabular}
}
\caption{F1 Scores for the BUCC 2020 Corpus Mining Challenge and the amount of parallel data used by that model for that language pair. Our method gets a lower score than the SoTA. However, when the extracted false positives were manually inspected, we found that most were viable sentence pairs. The issue with the BUCC dataset has been discussed in previous work in \citet{Reimers2020MakingMS}. We analyzed the reference sentences from the news dataset and observed that our method gave the correct parallel the highest score with close to 100\% accuracy.}
\label{table:bucc}
\end{table*}

The MultiQE Multitask model for scoring on the other hand, is trained on the MLQE with two variations. The first model is trained on the training set from MLQE datasets as before (section \ref{multitasktraining}), and the second model, which we will call MultiQE Multitask + DA(Data Augmentation), is trained with augmenting the dataset similar to our method for contrastive learning here, but instead of having labels 0 and 1, as in $Y_F$, here we keep the original  quality scores as the label set  
and give the negative samples a quality score of 0 and once again train our model in a multitask learning framework with the STS and MNLI data until convergence.

\subsubsection{Corpus Mining Inference}
TATOEBA has an equal number of sentences in both languages and we know that every sentence has a pair; the goal is to find the best sentence for each input. The test sets are reasonably small, so we directly use the scoring model to create the score matrix and pick the hypothesis with the highest score for each reference.

Because the BUCC filtering task has a more extensive test set, we do corpus mining in two stages. First, we use the trained filtration model to compute the sentence embedding for each sentence. Then we calculate the similarity matrix representing the similarities by multiplying the embedding vectors corresponding to every possible sentence pair. Then, for each sentence in the source and target domain, top-\textit{n} sentences are selected to be scored. The trained MultiQE Multitask model then scores these sentences. Then, for each sentence in the source and target domain, the best potential pair is selected by eliminating sentences whose scores are below a threshold. The QE scores that the MultiQE Multitask model provides range from 0-1 and the threshold score is determined similar to \citet{Reimers2020MakingMS} as the score that gives the best F1 score on the train set. The sentence selection is made in both directions and the intersection of the forward and the backward set is selected as the final filtered set. 
\begin{table*}\centering
\ra{1}
\resizebox{1\textwidth}{!}{%
\begin{tabular}{L{8cm}L{8cm}L{8cm}}\toprule
\textbf{German} & \textbf{English} & \textbf{Translation(Google Translate)} \\ 
     \midrule
    Nach dem Ende des Krieges erholte sich die Stadt aber rasch und wuchs beständig weiter.&	Following the end of the war the city continued to expand.	& After the end of the war, the city recovered quickly and steadily continued.  \\ 
     \hline
     Während einer Pestepidemie im Jahr 1541 starben rund 180 Personen, ein Viertel der Bevölkerung.&	During an epidemic of the plague in 1541 around 180 people died, a total of one fourth of the town’s residents.	& During a Pestepidemie in 1541, around 180 people died, a quarter of the population.\\ 
     \hline
     Eine Arbeitslosenversicherung gab es bis dahin nur im Bundesstaat Wisconsin (eingeführt 1932, wirksam wurde sie ab 1934).&	Unemployment insurance in the United States originated in Wisconsin in 1932.	&There was unemployment insurance only in the state of Wisconsin (introduced in 1932, it was effective from 1934). \\ 
     \hline
     Mehrere Universitäten in den Niederlanden bieten Studiengänge an, die die deutsche Sprache und Kultur vermitteln sollen.&	At academic level, 20 universities offer Dutch studies in the United States.&	Several universities in the Netherlands offer courses that should convey the German language and culture.\\
     \hline
     Im Juli 1994 war er nach dem Tod des Staatschefs Kim Il-sung an der Organisation der Trauerfeierlichkeiten beteiligt.&	He was a member of the funeral committee for Kim Il-sung in 1994.&	In July 1994 he was involved in the organization of mourning ceremonies after the death of the head of state of State.  \\ 
     \hline
     Im Jahr 1965 wurden dann die bestehenden politischen Parteien aufgelöst und ein künstliches Zweiparteiensystem geschaffen, das als „relative Demokratie“ bezeichnet wurde.&	Instead, in 1965, the government banned all existing political parties and created a two-party system.&	In 1965, the existing political parties were dissolved and created an artificial two-party system designated "relative democracy".\\
     \hline
     Am 22. Juni 1940 war der Waffenstillstand Hitlerdeutschlands mit dem besiegten Frankreich (de facto eine Kapitulation) unterschrieben worden. &	France was defeated and had to sign an armistice with Nazi Germany on June 22, 1940.&	On 22 June 1940, the ceasefire of Hitler Germans had been signed with defeated France (de facto a surrender). \\
     \hline
     Das Jahr 2004 wurde von den Vereinten Nationen zum "Reisjahr" erklärt.&	On December 16, 2002, the UN General Assembly declared the year 2004 the International Year of Rice.&	The year 2004 was explained by the United Nations on the "rice year".\\
     \hline
     Der Durchschnitt eines Haushalts bestand aus 3,55 Personen und die durchschnittliche Familie aus 3,54 Personen. &	The average household size was 4.05 and the average family size was 4.32. &	The average of a household consisted of 3.55 people and the average family of 3.54 people.\\
\bottomrule
\end{tabular}
}
\caption{A Random selection of false-negative pairs that the MultiQE Multitask + DA extracted from the BUCC de-en task. We can clearly see that while these sentences are labeled as negatives, they are actually meaningful parallel sentences supporting the existing arguments in the literature regarding the BUCC dataset. }
\label{table:manual_eval}
\end{table*}
\subsection{Experiments}
In table \ref{table:tatoeba} we show that our proposed method of multitask training and data augmentation is extremely effective in improving the robustness of QE models. We obtain an average performance increase of $0.80$ in accuracy compared to the SoTA QE method. We compare our method with Transquest \cite{transquest:2020} because both methods use XLM-R as the backbone and train on the exact same QE data. Our method performs comparably or better than extremely high resource methods like LASER \cite{Artetxe2019MassivelyMS} and Knowledge Distillation \cite{Reimers2020MakingMS} that require a lot more parallel data. Hence, these results become significant if we consider them together with the amount of parallel data used to train these models, which can be found in the same table.

The results are similar for the BUCC challenge (Table \ref{table:bucc}), where our method achieves comparable scores to SoTA methods that are trained on more than \emph{thousand} times the data. We can claim comparability because the F1 score in the BUCC task needs to be understood with a grain of salt. In Table \ref{table:manual_eval} we give some random examples of false negatives that are included in our model's selection of parallel sentences. As we can see, many of these sentences are as good parallels as the gold label set. As we have mentioned in Section \ref{sect:pcm}, the BUCC task injects news commentary data into Wikipedia and expects any method to only extract the injected data. This implicitly assumes that there are no correct parallel sentences within Wikipedia. Hence, the error our model displays is not failing to find correct parallels for hypothesis sentences but finding parallels within the Wikipedia corpus. We have manually analyzed 200 sentences and found that 155 of them can actually be considered good parallels. This issue has been discussed in previous work as well \cite{reimers-2019-sentence-bert,Jones2021MajorityVW}.

\section{Discussion}
We show that semantic similarity and linguistic inference improve QE performance. We test for significance and show that our methods outperform SoTA QE methods(Table \ref{table:corr}). 

This intuition that pretraining with related tasks, especially with STS and NLI, is helpful for evaluating translations is in line with background and findings from \citet{bleurt} and \citet{kanenubia}. Moreover, QE benefiting from monolingual data shows that XLM-R can utilize the labels in monolingual datasets to make better inferences in a cross-lingual task. This is most likely because it is already a cross-lingual language model. 

Additionally, we show that multitask training for QE can improve the robustness of the model. We demonstrate \textbf{improvements in accuracy around 0.50 in the TATOEBA experiments} (Table \ref{table:tatoeba}) over other SoTA QE model. The robustness in the corpus mining task can be attributed to embedding information learned from the NLI and STS tasks and the distribution of these datasets, where we have negative samples that allow our model to learn to eliminate unrelated sentences. 

We show that SoTA QE models yield unexpectedly poor performance in a PCM setting(Table \ref{table:tatoeba}). This is mainly due to how the QE data is created. The dataset only consists of sentence pairs generated by NMT models, which are translations of each other. They are either good or bad translations in a grammatical sense, but there are no non-translations, i.e., sentence pairs that are grammatical but are just unrelated. Hence a model trained on QE data only focuses on fluency and grammaticality and may unexpectedly rewards basic sentences where NMT models do not make mistakes because they always have a quality score of 1. To remedy this problem, we used negative data augmentation to "balance" the dataset and showed that this improves the performance on PCM, resulting in \textbf{an additional 0.30 increase and a total of 0.80 increase in accuracy}(Table \ref{table:tatoeba}). 

Our QE models process input sentences as pairs, bringing up the computational cost problem. Solving this with the sentence filtration model we train using contrastive loss enables the use of our QE method in large filtration tasks. Making it a good low resource corpus mining method that can achieve on par results with SoTA methods (Tables \ref{table:tatoeba} and \ref{table:bucc}). The importance of this contribution is amplified when we consider that our method is trained only using 7K parallel sentences compared to other PCM methods, which are trained on the order of millions of sentences.

Throughout our experiments, we keep low resource limitations in mind. While we acknowledge that collecting more data across different families of languages is an option to scale methods to low resource languages. We argue that exploring the improved usage of less data with \textit{better} labels promises another important avenue to make useful methods like QE or PCM available in low resource languages.

\section{Future Work}
To further our work, we plan to explore contrastive loss fine-tuning with self-supervision to improve QE performance planning and further reduce the need for labels. Self-supervised learning is an exciting way of forcing a neural language evaluator to abstain from certain mistakes. This approach can force invariance or target to reduce certain types of errors. The nature of the information attained by the network is primarily dependent on the negative sample generation process. 

Another interesting avenue to explore is using QE in active learning for machine translation as a scheduling or training signal. 

\section*{Acknowledgments}
This work is supported in part by the U.S. NSF grant 1838193 and DARPA HR001118S0044 (the LwLL program). The U.S. Government is authorized to reproduce and distribute
reprints for Governmental purposes. The views and conclusions contained in this publication are those of the authors and should not be interpreted as representing official policies or endorsements of DARPA and the U.S. Government.

We also want to thank Ekin Akyurek for his support in writing and structuring the paper. His insightful suggestions are much appreciated. 

\bibliography{anthology,custom}
\bibliographystyle{acl_natbib}

\appendix

\section{Appendix A}
\label{sec:appendix}

In this part, we will look over the distribution of QE scores for the language pairs in the MLQE dataset. The MLQE dataset is constructed - source sentences from Wikipedia are selected and translated using NMT methods; expert translators then score the outputs following FLORES methodology. This in turn had a few critical effects. As we mentioned in the paper, the first is that no sentence has been paired with grammatically correct sentences but is not related to that sentence. Every hypothesis sentence is intended to be a reasonable translation of that source sentence.  

The second outcome we have observed is that the QE model essentially adapts to the errors of the NMT model. The QE models only encounter low scores in the type of errors that NMT models are prone to making. Vice-versa, they see high scores, generally 1s in basic sentences where NMTs never make errors. This creates a specific type of error in QE performance where sentences that are easy to translate or need no virtual translation besides a few dictionary operations always receive high scores from the QE model no matter the source sentence, e.g., "June 10 1981" and "10. Juni 1981" from en-de. These types of elementary sentences were the highest scoring candidates for sometimes thousands of sentences in the BUCC dataset, constantly receiving scores close to 1. 

The distribution of the scores is mostly consistent with our findings. We only see that the low resource language pairs seem to have a better distribution across the board. While this seems to be a better case, it is not because the problem we mentioned does not exist, but because the NMT models that do the translation for these low resource languages perform worse.

\begin{figure}[h]
    \centering
    \includegraphics[scale=0.09]{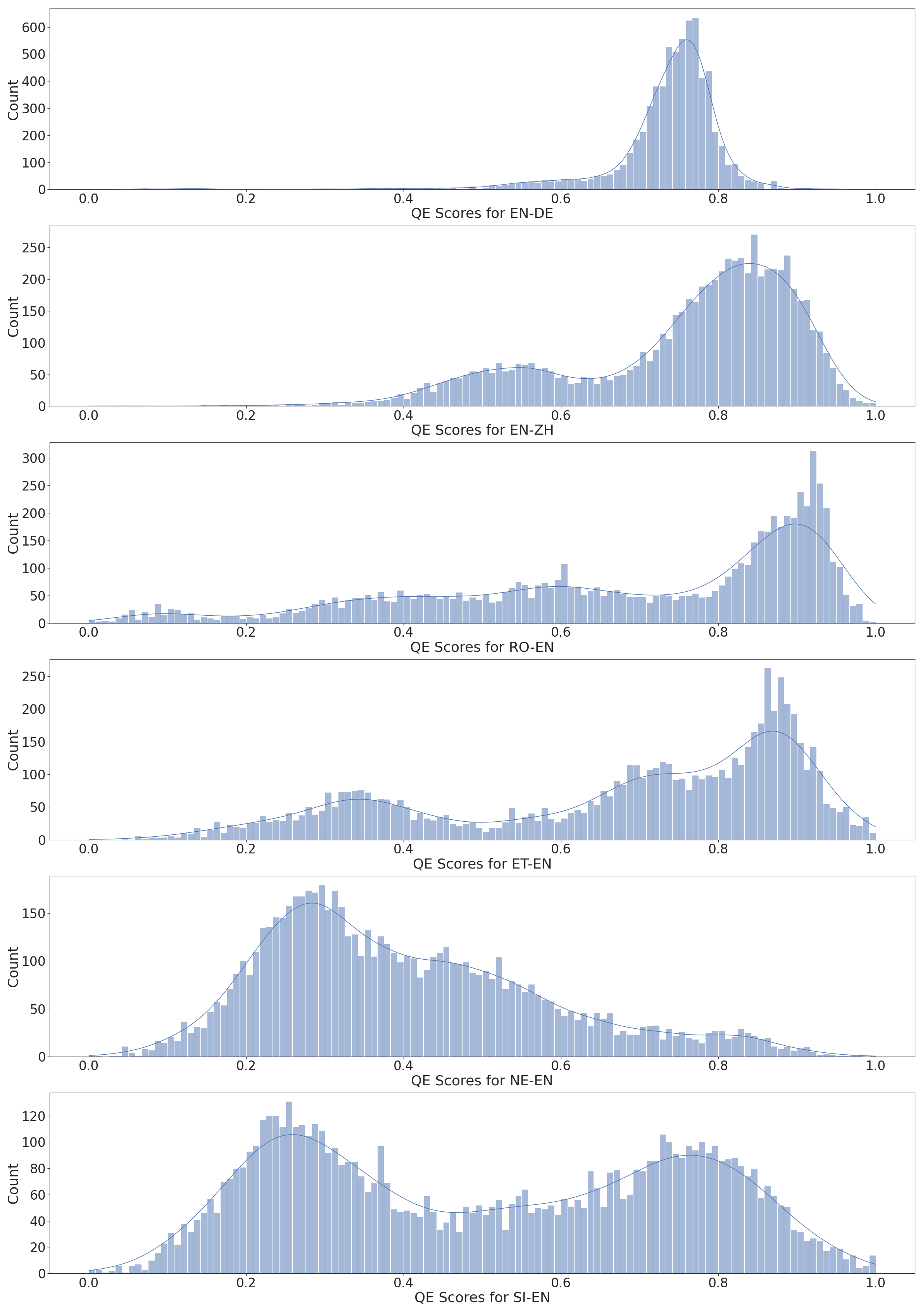}
    \caption{Distribution of QE scores from the MLQE datasets train split for all 6 language pairs }
    \label{fig:qe_dist}
\end{figure}

\end{document}